\documentclass{article} 
\usepackage[most]{tcolorbox}
\usepackage{colm2024_conference}
\lstdefinestyle{prompt_json}{
  basicstyle=\ttfamily,
  keywordstyle=\color{blue},
  stringstyle=\color{orange},
  commentstyle=\color{green},
  frame=single,
  rulecolor=\color{black},\textit{}
  breakatwhitespace=false,
  breaklines=true,
  captionpos=b,
  keepspaces=true,
  showspaces=false,
  showstringspaces=false,
  showtabs=false,
  tabsize=2
}

\tcbset{
  aibox/.style={
    top=10pt,
    colback=white,
    colframe=black,
    colbacktitle=black,
    enhanced,
    center,
    attach boxed title to top left={yshift=-0.1in,xshift=0.15in},
    boxed title style={boxrule=0pt,colframe=white,},
  }
}
\newtcolorbox{AIbox}[2][]{aibox, title=#2,#1}
\usepackage{microtype}
\usepackage{hyperref}
\usepackage{url}
\usepackage{alltt}
\usepackage{booktabs}
\usepackage{graphicx}
\usepackage{diagbox}
\usepackage{hyperref}
\usepackage{url}
\usepackage{tabularx}
\usepackage{graphicx}
\usepackage{booktabs}
\usepackage{subcaption}
\usepackage{multirow} 
\usepackage{pifont}
\usepackage{listings}
\usepackage[most]{tcolorbox}
\usepackage{colortbl}
\usepackage{hyperref}
\usepackage{marvosym}

\usepackage{listings}
\definecolor{darkblue}{rgb}{0, 0, 0.5}
\hypersetup{colorlinks=true, citecolor=darkblue, linkcolor=darkblue, urlcolor=darkblue}

\begin{document}
\title{CompassJudger-1: All-in-one Judge Model Helps Model Evaluation and Evolution}

\colmfinalcopy

\author{Maosong Cao, Alexander Lam, Haodong Duan, Hongwei Liu \\ {\bf Songyang Zhang$^{*}$, Kai Chen$^*$ }\\
Shanghai AI Laboratory \\
\texttt{opencompass@pjlab.org.cn}\\
\noindent {$^*$ corresponding authors}
}
\maketitle

%


\begin{abstract}

Efficient and accurate evaluation is crucial for the continuous improvement of large language models (LLMs).
Among various assessment methods, 
subjective evaluation has garnered significant attention due to its superior alignment with real-world usage scenarios and human preferences.
However, human-based evaluations are costly and lack reproducibility, 
making precise automated evaluators (judgers) vital in this process.
In this report, 
we introduce \textbf{CompassJudger-1}, 
the first open-source \textbf{all-in-one} judge LLM. 
CompassJudger-1 is a general-purpose LLM that demonstrates remarkable versatility. 
It is capable of:
1. Performing unitary scoring and two-model comparisons as a reward model; 
2. Conducting evaluations according to specified formats; 
3. Generating critiques;
4. Executing diverse tasks like a general LLM. 
To assess the evaluation capabilities of different judge models under a unified setting, 
we have also established \textbf{JudgerBench}, 
a new benchmark that encompasses various subjective evaluation tasks and covers a wide range of topics. 
CompassJudger-1 offers a comprehensive solution for various evaluation tasks while maintaining the flexibility to adapt to diverse requirements. 
Both CompassJudger and JudgerBench are released and available to the research community at \href{https://github.com/open-compass/CompassJudger}{CompassJudger}. 
We believe that by open-sourcing these tools, 
we can foster collaboration and accelerate progress in LLM evaluation methodologies.

\end{abstract}

\section{Introduction}

The evaluation of large language models (LLMs) typically falls into two major categories: objective evaluation and subjective evaluation. 
Objective evaluation evaluates LLMs using questions with ground-truth answers (multiple-choice questions, fill-in-the-blank questions, \textit{etc.}). 
Given LLM responses and answers, 
rule-based approaches are adopted to check if the responses are correct or not. 
Subjective evaluation, on the other hand, 
often requires a more comprehensive assessment of models' responses from multiple perspectives such as helpfulness, honesty, or creativity~\citep{liu2023alignBench}. 
Most questions in subjective evaluation do not have a ground-truth answer, 
thus requiring human judgement for response quality assessment. 

Despite the effectiveness, 
human-based subjective evaluation is time-consuming, 
labor-intensive, as well as hard to reproduce.
Therefore, in early stages of subjective evaluation, 
using the best-performing model (such as GPT4~\citep{achiam2023gpt}) to evaluate the responses of LLMs became the most adopted approach, which has been applied to multiple subjective evaluation benchmarks including AlpacaEval~\citep{dubois2024length}, ArenaHard~\citep{li2024crowdsourced}, WildBench~\citep{lin2024wildBench}, and AlignBench~\citep{liu2023alignBench}. 
However, those judge models with stronger critique capabilities are often commercial APIs with limited transparency 
and charge per use. 
The cost can be prohibitive for research studies when evaluations are frequently conducted.

Recent efforts in the research community have focused on 
developing open-source judge models for subjective evaluation of large language models (LLMs).
For example, CritiqueLLM~\citep{ke2023critiquellm} has been developed to evaluate AlignBench instead of GPT-4. 
Works like Auto-J~\citep{li2023generative} and Prometheus~\citep{kim2024biggen} also explore different training paradigms for judge models. 
However, most open-source judge LLMs are constrained by their specific training tasks and can only adhere to certain formats.
This limitation restricts their applicability across multiple subjective benchmarks simultaneously.
We posit that, 
rather than developing LLMs solely for judging purposes, 
it is crucial for a judge LLM to excel in both judging tasks and general tasks to achieve good generalization capability, similar to GPT-4.
Therefore, the development of an all-in-one model with strong judge capabilities is paramount. 
Such a model should demonstrate:
1. Robust generalization abilities;
2. Applicability to diverse subjective evaluation settings;
3. Adaptability to emerging evaluation tasks.

In addition, evaluating the effectiveness of judge models on subjective evaluation tasks is crucial.
However, among the numerous existing LLM benchmarks,
very few are specifically designed for assessing judge models or reward models. 
RewardBench~\citep{lambert2024rewardBench} represents an initial attempt in this direction, 
comprising approximately 3,000 questions with corresponding answers from two models and annotated ground truth preferences from humans. 
This benchmark aims to test whether judge models or reward models can accurately identify relative response quality.
Despite being a commendable first step, 
the scenarios covered by RewardBench are relatively simplified.  
It also lacks an examination of the model's critique capabilities and the ability to generate formatted judgement.
Given these constraints, 
there is a clear need for a more comprehensive benchmark to assess judgment capabilities.



In response to the aforementioned challenges,
we conduct extensive investigations on two key components: all-in-one judger training and judge model evaluation. 
High-quality data is paramount to the effective training of models, therefore, we first explore the data recipe for training CompassJudger-1, which includes various data sources such as pair-wise data used for reward model tuning, 
open-source critique data, 
and self-collected data. 
We further develop several data filtering and sampling strategies for the training process. 
Additionally, we also investigate the influence of the general SFT (G-SFT) data on training a judge model.
We also introduce the JudgerBench in this work, 
which includes the bi-lingual realistic human annotation from the LLM arena (ChatBot Arena~\citep{chiang2024chatbot} and CompassArena\footnote{LLM Arena released by OpenCompass: \href{https://arena.opencompass.org.cn/}{https://arena.opencompass.org.cn/}}), as well as GPT annotations on the typical subjective Benchmarks.
The related code and models are released in \href{https://github.com/open-compass/CompassJudger}{https://github.com/open-compass/CompassJudger}.

\section{CompassJudger-1}
In this section, we first introduce the training data used to build CompassJudger-1 in Sec.~\ref{sec:data_collect}, followed by the details of training strategies in Sec.~\ref{sec:training}

\subsection{Data Collection}\label{sec:data_collect}
\begin{figure*}[!t]
  \centering  
\includegraphics[width=\textwidth]{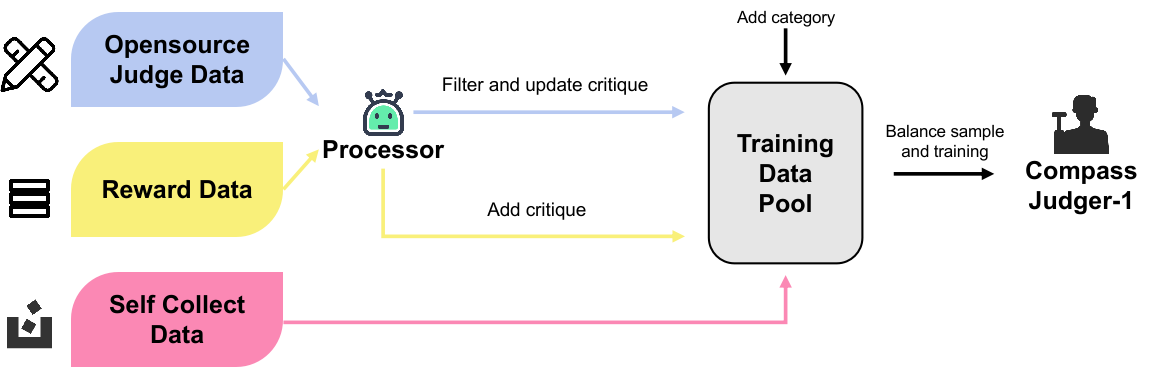}
  \caption{\textbf{Training Data Collection of CompassJudger-1.}}
  \label{fig:data}
\end{figure*}

High-quality data is essential for training an effective model, and the judge model is no exception. Therefore, we focus on data collection and the composition of the training dataset. Our training data is primarily divided into three categories: publicly available judge data, self-collected subjective evaluation data, and reward data used for training reward models.\par

We present the relevant information of all the data in Table~\ref{tab:data}.
The table clearly shows that the number of entries in the training data pool from different sources can vary greatly, making it highly imbalanced. Due to the unique nature of judge data, the output format from the same dataset is often uniform. This can cause the model's response patterns to become rigid, which is a key reason why most judge models struggle with out-of-domain evaluation. Therefore, appropriate sampling strategy is necessary when incorporating the data into the final training set.

\begin{table*}[!t]
\centering
\caption{\textbf{Training Data Construction of CompassJudger-1}, \textbf{Pointwise} indicates that the data contains only one model's response along with the score given by the Judge model/Reward model. \textbf{Pairwise} indicates that the data includes responses from two models and the comparison result given by the Judge model/Reward model. \textbf{Generative} indicates that the data includes the Judge results as well as the reasoning process of the Judge. The number of each dataset refers to the number of candidates in the Training Data pool, not the final amount of training data.} 
\vspace{-2mm}
\label{tab:data}
\resizebox{\linewidth}{!}{
\begin{tabular}{lllll}
\midrule
Attribution  & Dataset Name & Data Format & Number&Language \\
\midrule
Open-source Judge Data &AlpacaFarm~\citep{dubois2024alpacafarm} &Pairwise&39k&EN\\
& Auto-J~\citep{li2023generative}& Pointwise, Pairwise, Generative&9k&ZH, EN\\
&PandaLM~\citep{wang2023pandalm}& Pairwise&287k&EN\\
&JudgeLM~\citep{zhu2023judgelm}& Pointwise&100k&EN\\
&LLM-Eval2~\citep{LLMEval2} &Pointwise, Generative&10k&ZH\\
&CritiqueBench~\citep{lan2024criticBench} &Generative&1k&EN\\
&UltraFeedback~\citep{cui2023ultrafeedback}&Pointwise, Generative&380k&EN\\
\midrule
Open-source Reward Data& OffsetBias~\citep{park2024offsetbias}  &Pairwise&8k&EN\\
&Hendrydong~\citep{dong2024rlhf}&Pairwise&700k&EN\\
&SkyWorker~\citep{skyworkcritic2024}&Pairwise&80k&EN\\
&Airoboros&Pairwise&36k&EN\\
&Anthropic&Pairwise&161k&EN\\
&PKU Alignment&Pairwise&82k&EN\\

\midrule
Self Collect Judge Data& CJ-Judge-Data-v1 & Pointwise, Pairwise, Generative& 60k&ZH, EN\\
\midrule
Self Collect Reward Data&Math Code Preference& Pairwise & 11k&EN\\
&Chinese Math& Pairwise& 76k &ZH\\
&LengthControl&Pairwise& 0.6k& EN\\
&Language Match&Pairwise& 0.5k&ZH, EN\\
\hline
\end{tabular}}
\end{table*}

Additionally, the proportion of generative data in the overall data pool is relatively small, with most data containing \textbf{only judgment results rather than the reasoning behind them}. Therefore, to strengthen the model’s critique capabilities, it is essential to expand the generative critique portion of the existing data. To address this, we introduce the following strategies for datasets from various sources.
The specific data processing procedures are also illustrated in Figure~\ref{fig:data}.



\paragraph{Public Judge Data:} 
We collect public available judge data from below datasets: PandaLM~\citep{wang2023pandalm}, JudgeLM~\citep{zhu2023judgelm}, AutoJ~\citep{li2023generative}, AlpacaFarm~\citep{dubois2024alpacafarm}, CritiqueBench~\citep{lan2024criticBench}, Ultrafeedback~\citep{cui2023ultrafeedback}, etc. Then, we introduce three methods to improve the training data quality of them.\par
\noindent$\bullet$ \textbf{Re-evaluate.} We start by updating outdated judge data, which uses older models like ChatGPT for judging and has lost relevance. 
Therefore, we use the more powerful Qwen-2.5-72B~\citep{yang2024qwen2}, to re-evaluate the original tasks and generate new training data.\par
\noindent$\bullet$ \textbf{Data Categorization.} We further categorize the data to facilitate subsequent sampling. The question components are extracted, and each question is assigned to a category using LLMs with specifically designed prompts. Detailed prompts and category labels can be found in the \ref{appendix:prompt1}.\par 
\noindent$\bullet$ \textbf{Balance Sampling}.
During the training of general models~\citep{cai2024internlm2}, we achieved better results when the various class labels in the training data maintain a certain level of balance. Thus to ensure category balance, we applied category-balanced sampling to the judge data. Using the class labels obtained from the previous process, we performed undersampling on categories with ample data and oversampling on categories with insufficient data, ultimately maintaining a balance across all categories in the dataset.

\paragraph{Self-collect Data} 

During the iterative training of InternLM2.5~\citep{cai2024internlm2}, we conducted extensive evaluation on our internal subjective test sets. We collected this internal data, which covers a wide range of question types, including pointwise, pairwise, and single-turn dialogue evaluations, and multi-turn dialogue evaluations. Since the judgments are made by models with strong evaluation capabilities, they are directly added to the training data pool after balancing the quantities, resulting in approximately 20k entries in the final trainingset.

\paragraph{Reward Data} 
Our early experiments indicates that excessive reward data, which lacked a critique process and contained only judge results, led to model collapse, necessitating strict limitations on its quantity to maintain balanced training data. To enhance the judge model's critique capabilities, particularly for reasoning and mathematics problems, we reformat questions and implemented a processing system to generate detailed critique processes, ultimately curating approximately 500,000 entries of high-quality data. The prompts for obtaining critiques are detailed\ in the ~\ref{appendix:prompt2}.
\par

\subsection{Training Strategy and Ablation Study}\label{sec:training}
In this section, we present our training configuration and discuss our ablation study on different data sources.

\paragraph{Training Configuration}


Based on the processed training data pool, we conduct experiments with varying data ratios. We employ Xtuner~\citep{2023xtuner} as our training framework and, through extensive experimentation, determine that an epoch of 2 and a learning rate of 2e-5 are optimal parameters. We perform ablation experiments specifically on the proportions of reward data and general SFT data.\par


\paragraph{Ablation Study on Reward Data}
We first examine the proportion of the reward data. As the pairwise evaluation tasks in reward data only require the categorical annotation, we can easily collect a large number of data from the public community.
While incorporating reward data benefits the training of an all-in-one judge model, our experiments reveal that excessive reward data can lead to model overfitting, resulting in simplistic outputs resembling a reward model and compromising the ability to perform complex critique tasks. Through experimentation (See in Table~\ref{tab:rewardablation}), we determine that the optimal proportion of reward data lies between 50\% and 70\%, enabling the model to achieve strong judging capabilities while maintaining generalizability.\par

\begin{table*}[!t]
\centering
\caption{\textbf{Ablation Study About the Proportion of Reward Data.}}%

\vspace{-2mm}
\label{tab:rewardablation}
\resizebox{\linewidth}{!}{
\begin{tabular}{l|c|ccc}
\midrule
Models& \textbf{Proportion of Reward Data}& \textbf{RewardBench}&  \textbf{JudgerBench}& \textbf{Average}\\
\midrule

CompassJudger-1-7B & 25\% & 0.810	& 0.633    &0.722\\
                   & 33\% & 0.812	 & 0.646	&0.729\\
                   & 50\% & 0.823	 & 0.665	&0.744\\
                   & \textbf{66\%} & 0.831	 & \textbf{0.697}    &\textbf{0.764}\\
                   & 75\% & 0.833	 & 0.612	&0.723\\
                   & 83\% & \textbf{0.834}	 & 0.438	&0.636\\
\hline
\end{tabular}}
\end{table*}
\begin{table*}[!thb]
\centering
\caption{\textbf{Ablation Study of General SFT Data.} "Judge Average" refers to the evaluation score that encompasses the judging capabilities of both RewardBench and JudgerBench, while "Subjective Average" is the evaluation score on several subjective datasets listed in the table. The relevant evaluation results are obtained using OpenCompass~\citep{2023opencompass}. All results from the corresponding datasets have been normalized to percentages.}%

\vspace{-2mm}
\label{tab:subeval}
\resizebox{\linewidth}{!}{
\begin{tabular}{l|c|ccccc}
\midrule
Models& \textbf{Judge Average}& \textbf{AlignBench}&  \textbf{ArenaHard}& \textbf{Fofo}& \textbf{WildBench}& \textbf{Sub. Average}\\
\midrule

CJ-1-7B-w/o G-SFT & 0.693 & 0.590	 & 0.487	 & 0.750	 & -0.071	 & 0.490\\
CJ-1-7B-w. G-SFT & 0.697	& 0.624	  & 0.562	& 0.740	 & 0.015	 &0.528  \\
\hline
\end{tabular}}
\end{table*}
\paragraph{Influence of General SFT Data }
We then investigate the impact of general SFT data proportions. Since our goal is to create a powerful all-in-one model with judging capabilities rather than a model specific to particular datasets, incorporating general SFT (G-SFT) training data is essential for maintaining generalizability. Our G-SFT data comes from the internal training data for SFT. Our experiments show that
an appropriate amount not only preserves the model's general capabilities but also enhances its judging performance. We present these results in Table \ref{tab:subeval}.\par


\paragraph{Data Recipes}
Based on our findings, we establish the optimal training data ratio as: 
\begin{equation}
    \texttt{\{critique data : reward data : general SFT data = 1 : 3 : 1\}}
\end{equation}
 Our final training dataset comprises approximately 900k entries. We select the recently released open-source Qwen2.5 series~\citep{yang2024qwen2} as the foundation models for our judge SFT training. We detail the related results in the following sections.

\section{JudgerBench}
\label{sec2}
In this section, we introduce JudgerBench, a specialized evaluation dataset designed to evaluate the judge models. To replicate realistic judge model application scenarios, JudgerBench incorporates two distinct types of annotations: \textbf{human annotations} for the arena part and \textbf{LLM annotations} for the benchmark component. We detail the construction process of these components in Sec.~\ref{subsec:construction} and present CompassJudger's performance on JudgerBench in Sec.~\ref{subsec:Bench_results}.


\subsection{JudgerBench Construction}\label{subsec:construction}
JudgerBench consists of two primary components: the \textbf{Arena} component (denoted as JDB-A) and the \textbf{Benchmark} component (denoted as JDB-B). JDB-A, similar to RewardBench, focuses on alignment with human preferences and requires only simple judge outputs such as \texttt{[[A]]} or \texttt{[[B]]}. JDB-B, on the other hand, assesses the model's critique capabilities and its ability to provide judgments following specific formats.\par

\paragraph{JudgerBench A (Arena Part)}\quad\par
\noindent$\bullet$ \textbf{Data Source.} JDB-A consists of two sections: English and Chinese. The English section is derived from the released data of Chatbot Arena~\citep{chiang2024chatbot}, while the Chinese section comes from the collected data of CompassArena~\citep{2023opencompass}. These data are the results of human voting, with each entry containing a question and the corresponding responses from two models, along with the vote on which model is the winner. Both Chinese and English include single-turn and multi-turn dialogue data, with approximately 500 single-turn dialogues and 250 multi-turn dialogues, totaling around 1500 pieces of data. We introduced in next paragraph how we screen and obtain these data.
 \par
\noindent$\bullet$ \textbf{Screening Process.} The 1,500 data points in JDB-A were obtained through the following screening process: We first performed unsupervised clustering (specifically, k-means with k set to 50 in our implementation) on all the data to get rough categories. Then, the Processor model (Qwen2.5-72B) summarized specific category names based on typical cases within these categories (detailed category names can be found in the results of Table~\ref{tab:detailA_EN}). After obtaining the top 10 category names summarized by the processor, we used the processor to process each data point individually, assigning them to their respective categories. Following the detailed categorization, we conducted manual screening to ensure the correctness of the judging data.

In addition, to give JDB-A a difficulty grading system, besides setting categories and handling single-turn and multi-turn dialogues, we also performed difficulty grading based on the length of model responses. According to the experience of subjective evaluation~\citep{dubois2024length}, Judgers often exhibit a length bias, meaning they tend to favor longer responses when the quality of responses from two models is similar. To avoid fitting this length bias of the Judgers, we selected a batch of shorter responses that were marked as winners to serve as difficult data, thus increasing the difficulty of JDB-A.\par

\paragraph{JudgerBench B (Benchmark Part)} \quad\par

\noindent$\bullet$ \textbf{Construction Methods.} JDB-B primarily includes four datasets (AlignBench~\citep{liu2023alignBench}, ArenaHard~\citep{li2024crowdsourced}, FoFo~\citep{xia2024fofo}, and WildBench~\citep{lin2024wildBench}), which are very commonly used in subjective evaluations, covering different subjective scenarios (such as daily chat, instruction following), different evaluation methods (such as scoring, head-to-head competition), different languages (Chinese and English), and single and multi-turn dialogues. We detail the relevant properties of these datasets in Table~\ref{tab:subintro}. For these four subjective datasets, we sampled 100 questions from each dataset according to their respective subcategories, totaling 400 questions. Then, we used the top 10 closely ranked models from the OpenCompass leaderboard\footnote{We use the data of 202407 version.} to obtain their responses to these 400 questions (specific model information can be found in the data json), thus acquiring a total of 4000 QA pairs, note that these 10 models have very similar and high capabilities, which also demonstrates the difficulty of judging with JDB-B. We then used GPT-4o\footnote{We use \texttt{gpt-4o-2024-08-06} as default if not specified.} to judge these pairs, using this judgment as a benchmark to check whether the judgment results of other models align with those of GPT-4o. \par

\begin{table*}[!t]
\centering
\caption{\textbf{Detailed Introduction of Subjective Evaluation Datasets in JDB-B} The official FoFo dataset includes only English, and we created the Chinese portion. Additionally, due to the outdated references in AlignBench, we changed its evaluation method from Pointwise to Pairwise.}
\vspace{-2mm}
\label{tab:subintro}
\resizebox{\linewidth}{!}{
\begin{tabular}{lllll}
\midrule
Dataset Name  & Data Format & Turns & Scenario Label &Language \\
\midrule
AlignBench &  Pairwise &  Single Turn & Daily Chat, Chinese Culture & ZH\\
ArenaHard &Pairwise & Single Turn &Daily Chat, Reasoning, Math, Code& EN\\
FoFo&Pointwise& Single Turn&Instruction Following&ZH, EN\\
WildBench&Pairwise& Single Turn, Multi Turn&Daily Chat&EN\\
\hline
\end{tabular}}
\end{table*}

\begin{table*}[!t]
\centering
\caption{\textbf{Results on RewardBench and JudgerBench,} Which JDB-A means JudgerBench partA, JDB-B means JudgerBench partB. }

\vspace{-2mm}
\label{tab:total}
\resizebox{\linewidth}{!}{
\begin{tabular}{l|c|ccccc}
\midrule
Models& \textbf{RewardBench}& \textbf{JDB-A EN}& \textbf{JDB-A CN}& \textbf{JDB-B Acc}& \textbf{JDB-B Corr}& \textbf{JudgerBench}\\
\midrule

Qwen2.5-7B-Chat        &0.789 &0.567 &0.535 &0.590 &0.874 &0.641 \\
Qwen2-72B-Chat         &0.822 &0.588 &0.584 &0.625 &0.935 &0.683 \\
Qwen2.5-72B-Chat       &0.832 &0.615 &0.590 &0.681 &0.937 &0.706 \\
\midrule
GPT-4o-0806             &0.867 &0.664 &0.608 &1     &1     &0.818 \\
Skywork-llama3.1-8B    &0.890 &0.630 &0.605 &-     &-     &-     \\
Selftaught-llama3.1-70B &0.900 &0.443 &0.570 &0.598 &0.869 &0.620 \\
\midrule
CJ-1-1.5B              &0.724 &0.553 &0.527 &0.629 &0.905 &0.654 \\
CJ-1-7B                &0.831 &0.570 &0.583 &0.687 &0.948 &0.697 \\
CJ-1-14B               &0.842 &0.599 &0.615 &0.699 &0.959 &0.718 \\
CJ-1-32B               &0.854 &0.614 &0.612 &0.720 &0.963 &0.727 \\

\hline
\end{tabular}}
\end{table*}

\noindent$\bullet$ \textbf{Evaluation Metrics.} To facilitate the research, we adopt the GPT-4o's judgement as the reference ground-truth (though there may exist noise and errors). We calculate from two metrics for JDB-B, accuracy and correlation.
\begin{itemize}
    \item  The accuracy per question, i.e., whether each model's judgment on each question matches GPT-4o's judgment, which is a very stringent indicator, especially when dealing with datasets like WildBench that require five-category judgments (A++, A+, A=B, B+, B++). The accuracy rate of judge models tends to be lower. 
    \item  For the correction metric, we adopt \textbf{Pearson product-moment correlation coefficients}. We first obtain the scoring results of 10 reference models and then sort this result. If these average scores on whole dataset align closely with GPT-4o's scores, then even if there are discrepancies in the scoring on each question, it indicates that the Judge Model can reflect good judgment effects overall. 
\end{itemize}

\subsection{JudgerBench Results}\label{subsec:Bench_results}

\paragraph{Overall Results}\quad\par
We test baseline Chat models, current SOTA Judge Models, and our CompassJudger series on RewardBench and JudgerBench, reporting the overall results in Table \ref{tab:total}, with more detailed results presented in subsequent tables. Notably, many judge models failed to adhere to the prompt templates of the subjective datasets, leading to test failures, and we showcase one of these failure cases in Appendix~\ref{appendix:false}. 
From the table, it can be observed that our CompassJudger series outperforms all open-source models and achieves over 95\% of GPT-4o's judging capability in the relevance tests on JDB-B. While GPT-4o demonstrates high consistency with human evaluation results across different data domains, some models (such as Selftaught, skyworker), despite achieving high scores in one domain like RewardBench, lose a certain degree of generalization in other domains and do not possess good universal judging capabilities. Even though they are generative models, they are no longer able to follow instructions to evaluate common subjective datasets. In contrast, CompassJudger v1 achieved relatively balanced results on both RewardBench and JDB-A, and showed a significant lead on JDB-B.


\paragraph{RewardBench Results}\quad\par
We present the detailed results on RewardBench in Table~\ref{tab:detailreward}. For RewardBench, some existing Judge Models perform well (e.g. Skywork and Selftaught), even surpassing GPT-4o. However, upon closer inspection, the main gap is evident in the Chat Hard category. The number of questions in this category does not constitute a large proportion of the total questions on RewardBench. There is a possibility of over-training with respect to these Judge Models, and our CompassJudger series also shows improving scores in this category as the model size increases.
\begin{table*}[!t]
\centering
\caption{\textbf{Detailed Results on RewardBench.}}
\vspace{-2mm}
\label{tab:detailreward}
\resizebox{0.7\linewidth}{!}{
\begin{tabular}{l|ccccc}
\midrule
Models& \textbf{Chat}& \textbf{Chat Hard}& \textbf{Safety}& \textbf{Reasoning}& \textbf{Average}\\
\midrule

Qwen2.5-7B-Chat & 0.961 & 0.567	& 0.831	 & 0.797	 & 0.789	\\
Qwen2-72B-Chat& 0.955	& 0.640 & 0.843	 & 0.848	& 0.822	    \\
Qwen2.5-72B-Chat& 0.961	& 0.680 & 0.838	 & 0.850	& 0.832	  \\
\midrule
GPT-4o-0806& 0.961	& 0.761 & 0.881	 & 0.866	& 0.867	  \\
Skywork-llama3.1-8B& 0.936	& 0.814 & 0.911	 & 0.898	& 0.890	  \\
Selftaught-llama3.1-70B& 0.969	& 0.851 & 0.896	 & 0.884	  & 0.900   \\
\midrule
CJ-1-1.5B& 0.964 & 0.495 & 0.781 & 0.656 & 0.724	  \\
CJ-1-7B& 0.978	& 0.605 & 0.847	 & 0.895	& 0.831	 \\
CJ-1-14B& 0.975	& 0.623 & 0.845 & 0.925& 0.842	  \\
CJ-1-32B& 0.978	& 0.656 & 0.861	 & 0.922	& 0.854	  \\

\hline
\end{tabular}}
\end{table*}

\begin{table*}[!t]
\centering
\caption{\textbf{Detailed Results on JDB-A-EN.}}

\vspace{-2mm}
\label{tab:detailA_EN}
\resizebox{\linewidth}{!}{
\begin{tabular}{l|cccccccccc}
\midrule
Models& \textbf{Teaser}& \textbf{AI}& \textbf{Roleplay}& \textbf{Chat}& \textbf{Math}& \textbf{Reasoning}& \textbf{Creation}& \textbf{Code}& \textbf{Science}& \textbf{Humanities}\\
\midrule

Qwen2.5-7B-Chat         & 0.54 & 0.59 & 0.59 & 0.46 & 0.69 & 0.43 & 0.61 & 0.65 & 0.58 & 0.52\\
Qwen2-72B-Chat         & 0.63 & 0.59 & 0.54 & 0.49 & 0.62 & 0.64 & 0.60 & 0.74 & 0.51 & 0.52 \\
Qwen2.5-72B-Chat       & 0.68 & 0.57 & 0.57 & 0.47 & 0.78 & 0.64 & 0.58 & 0.75 & 0.61 & 0.52 \\ 
\midrule
GPT-4o-0806& 0.82 & 0.53 & 0.62 & 0.61 & 0.83 & 0.67 & 0.67 & 0.73 & 0.64 & 0.55 \\ 
Skywork-llama3.1-8B& 0.69 & 0.61 & 0.54 & 0.62 & 0.63 & 0.64 & 0.60 & 0.69 & 0.74 & 0.53 \\ 
Selftaught-llama3.1-70B& 0.47 & 0.45 & 0.47 & 0.37 & 0.45 & 0.43 & 0.36 & 0.58 & 0.48 & 0.36 \\ 
\midrule
CJ-1-1.5B              & 0.42 & 0.56 & 0.56 & 0.43 & 0.66 & 0.47 & 0.55 & 0.78 & 0.64 & 0.44 \\ 
CJ-1-7B                & 0.56 & 0.56 & 0.51 & 0.47 & 0.68 & 0.58 & 0.58 & 0.75 & 0.58 & 0.43 \\
CJ-1-14B               & 0.66 & 0.51 & 0.57 & 0.54 & 0.72 & 0.61 & 0.56 & 0.74 & 0.61 & 0.47 \\
CJ-1-32B               & 0.66 & 0.57 & 0.56 & 0.59 & 0.78 & 0.58 & 0.55 & 0.75 & 0.60 & 0.49 \\

\hline
\end{tabular}}
\end{table*}

\paragraph{JudgerBench A Results}\quad\par

The detailed results for JudgerBench A are presented in Tables ~\ref{tab:detailA_EN} and ~\ref{tab:detailA_CN}, showing that there are slight differences in the judging capabilities of various models in both Chinese and English domains. For instance, the English reasoning judging performance of Qwen2.5-7B-Chat is lower than its Chinese reasoning performance, and the English scores for humanities are also much lower than the Chinese scores for all models. On the other hand, GPT-4o leads in both Chinese and English across all domains, particularly in the Teaser category, where GPT-4o significantly outperforms other models, indicating its better understanding and discernment of human teasers. In the Chat category, the scores of the models are relatively close in both Chinese and English.
From the scoring situation of CompassJudger, its strengths are evident in Math, Reasoning, and Code, and these sub-domains are closely related to the model's judging capabilities. Therefore, the high judging performance of CompassJudger is also attributed to the enhanced reasoning and mathematical abilities in its training data, which also provides insights for the development of subsequent Judge Models and general-purpose models.


\begin{table*}[!t]
\centering
\caption{\textbf{Detailed Results on JDB-A-CN.}}

\vspace{-2mm}
\label{tab:detailA_CN}
\resizebox{\linewidth}{!}{
\begin{tabular}{l|cccccccccc}
\midrule
Models& \textbf{Teaser}& \textbf{AI}& \textbf{Roleplay}& \textbf{Chat}& \textbf{Math}& \textbf{Reasoning}& \textbf{Creation}& \textbf{Code}& \textbf{Science}& \textbf{Humanities}\\
\midrule

Qwen2.5-7B-Chat         & 0.46 & 0.58 & 0.36 & 0.45 & 0.70& 0.53 & 0.52 & 0.53 & 0.52 & 0.64 \\ 
Qwen2-72B-Chat         & 0.62 & 0.54 & 0.34 & 0.55 & 0.68 & 0.63 & 0.58 & 0.58 & 0.62 & 0.64 \\ 
Qwen2.5-72B-Chat       & 0.65 & 0.47 & 0.49 & 0.47 & 0.71 & 0.60& 0.57 & 0.58 & 0.69 & 0.60\\
\midrule
GPT-4o-0806& 0.77 & 0.56 & 0.51 & 0.53 & 0.67 & 0.66 & 0.63 & 0.58 & 0.62 & 0.58 \\ 
Skywork-llama3.1-8B& 0.62 & 0.58 & 0.58 & 0.59 & 0.63 & 0.58 & 0.60& 0.61 & 0.60& 0.61 \\ 
Selftaught-llama3.1-70B& 0.62 & 0.56 & 0.55 & 0.48 & 0.67 & 0.55 & 0.57 & 0.57 & 0.51 & 0.61 \\
\midrule
CJ-1-1.5B               & 0.54 & 0.58 & 0.38 & 0.38 & 0.62 & 0.63 & 0.54 & 0.52 & 0.55 & 0.54 \\ 
CJ-1-7B                & 0.62 & 0.54 & 0.41 & 0.58 & 0.70& 0.60& 0.59 & 0.56 & 0.59 & 0.60\\ 
CJ-1-14B               &  0.69 & 0.61 & 0.51 & 0.55 & 0.71 & 0.68 & 0.60& 0.58 & 0.61 & 0.65 \\
CJ-1-32B              & 0.69 & 0.58 & 0.53 & 0.52 & 0.71 & 0.53 & 0.60& 0.61 & 0.61 & 0.69 \\ 

\hline
\end{tabular}}
\end{table*}

\paragraph{JudgerBench B Results}\quad\par
On JudgerBenchB, we report the results based on the accuracy rate per question and the correlation results based on the overall model scores, shown in Tables~\ref{tab:detail_b_acc} and ~\ref{tab:detail_b_corr}, respectively. Many models, although they are generative Judge Models, are fundamentally unable to evaluate according to the instructions given by the dataset, and therefore we cannot report their scores on JudgerBenchB. This is also one of the original intentions behind our proposal for an All-in-one judge model.
From the tables, we can see that even though Selftaught-llama3.1-70B is a more generalizable Judge Model than Skywork-llama3.1-8B, the performance of this specialized Judge Model in subjective evaluations is even inferior to that of general models like the Qwen-chat series. This also indicates that the domains for training current Judge Models are too narrow, while CompassJudger maintains excellent generalizability and judging capabilities among them.

\begin{table*}[!t]
\centering
\caption{\textbf{Detailed Accuracy Results on JDB-B.}}

\vspace{-2mm}
\label{tab:detail_b_acc}
\resizebox{0.8\linewidth}{!}{
\begin{tabular}{l|ccccc}
\midrule
Models& \textbf{AlignBench}& \textbf{Fofo}& \textbf{WildBench}& \textbf{ArenaHard}& \textbf{Average}\\
\midrule

Qwen2.5-7B-Chat        & 0.777                    & 0.670               & 0.470                    & 0.444                   & 0.590 \\
Qwen2-72B-Chat        & 0.867                    & 0.692              & 0.564                   & 0.376                   & 0.625 \\
Qwen2.5-72B-Chat       & 0.878                    & 0.677              & 0.599                   & 0.570                    & 0.681 \\
Selftaught-llama3.1-70B& 0.755                    & 0.627              & 0.538                   & 0.472                   & 0.598 \\
\midrule
CJ-1-1.5B             & 0.822                    & 0.712              & 0.550                    & 0.430                    & 0.629 \\
CJ-1-7B                & 0.816                    & 0.783              & 0.564                   & 0.586                   & 0.687 \\
CJ-1-14B               & 0.839                    & 0.787              & 0.566                   & 0.602                   & 0.699 \\
CJ-1-32B              & 0.857                    & 0.806              & 0.596                   & 0.621                   & 0.720 \\

\hline
\end{tabular}}
\end{table*}

\begin{table*}[!hthb]
\centering
\caption{\textbf{Detailed Correlation Results on JDB-B.}}

\vspace{-2mm}
\label{tab:detail_b_corr}
\resizebox{0.8\linewidth}{!}{
\begin{tabular}{l|ccccc}
\midrule
Models& \textbf{AlignBench}& \textbf{Fofo}& \textbf{WildBench}& \textbf{ArenaHard}& \textbf{Average}\\
\midrule

Qwen2.5-7B-Chat        & 0.916                    & 0.681              & 0.967                   & 0.931                   & 0.874\\
Qwen2-72B-Chat         & 0.937                    & 0.889              & 0.976                   & 0.936                   & 0.935 \\
Qwen2.5-72B-Chat       & 0.964                    & 0.916             & 0.958                   & 0.912                   & 0.937\\
Selftaught-llama3.1-70B& 0.918                    & 0.667             & 0.950                    & 0.942                   & 0.869\\
\midrule
CJ-1-1.5B              & 0.928                    & 0.851              & 0.981                   & 0.858                   & 0.905 \\
CJ-1-7B                & 0.956                    & 0.936             & 0.970                    & 0.932                   & 0.948\\
CJ-1-14B               & 0.966                    & 0.956              & 0.965                   & 0.951                   & 0.959 \\
CJ-1-32B               & 0.973                    & 0.951             & 0.954                   & 0.975                   & 0.963\\

\hline
\end{tabular}}
\end{table*}

\section{Conclusion and Discussion}\label{sec4}
In this report, we propose the all-in-one Judge model, CompassJudger, and the JudgerBench for evaluating model judging capabilities. Our CompassJudger model has achieved the best results among open-source models on JudgerBench and is truly capable of replacing GPT-4o for evaluating common subjective datasets, greatly reducing the cost of subjective evaluations. In addition, we have many issues for further discussion.

\paragraph{How can the Judge Model assist in the iteration of models?}

The potential of a good judge model is not limited to just judging and critiquing capabilities; it can also help models iterate and evolve. It should point out the shortcomings of the model when answering questions and provide guidance, which is more conducive to the model improving its responses or achieving a more standardized answer—something that a regular reward model cannot accomplish. This has been validated in related experiments, and our internal experimental results are coming soon, which will demonstrate how we use the judge model to facilitate better iteration of the model.

\paragraph{If the model's judging capability is part of its general abilities, can judge training enhance the model's overall general capabilities?}

Just as GPT-4 can handle all judge tasks, we believe that judge capability is just a part of the model's general abilities, with a focus on reasoning and instruction following. Our experiments have also observed that good training in instruction following and reasoning abilities can significantly improve the model's judging capability. Conversely, relevant judge data can further enhance the model's reasoning and instruction following abilities. 

\section{Acknowledgement}

We would like to express our sincere gratitude to InternLM's post-training team for generously providing the data used in this research. Their contribution is essential to the success of this project. We also extend our thanks to Jiaye Ge for their invaluable support, guidance, and coordination throughout the project.

\newpage
\bibliography{colm2024_conference}
\bibliographystyle{colm2024_conference}

\newpage
\appendix
\section{Appendix}

\subsection{Prompt and category label for data categorization of public judge data.}\label{appendix:prompt1}

\begin{figure*}[!ht] 
\vspace{-5mm}
\begin{AIbox}{Prompt for categorization}
CATEGORY MAP = ["General Q\&A", "Tech Consulting", "Education Tutoring", "Healthcare", "Travel", "Finance \& Investment", "Legal Advice", "Psychological Counseling", "Entertainment Gossip", "Cuisine \& Cooking", "Home Improvement", "Auto Maintenance", "Video Games", "Sports \& Fitness", "Literature \& Art", "History \& Humanities", "Politics \& Current Events", "Religion \& Faith", "Parenting Education", "Pet Care", "Career Planning", "Shopping Recommendations", "Lifestyle Services", "Relationships \& Emotions", "Social Networking", "Programming \& Development", "Data Analysis", "Marketing", "Business Management", "Entrepreneurship Guidance", "Scientific Exploration", "Environmental Protection", "Other"]
\par\vspace{5pt}
CATEGORY PROMPT = I need to categorize the user's question. Below is a category map. Please help me categorize the user's question into one of these categories.\par
The category map is as follows: \{category map\}\par
[Start of User Question]\par
\{question\}\par
[End of User Question]\par
Please provide the category strictly in the following format and do not output any other statements:\par
[Start of Question Categorization]\par
xxxx\par
[End of Question Categorization]
\end{AIbox} 
\end{figure*}

\subsection{Prompt for getting critique of reward data.}\label{appendix:prompt2}

\begin{figure*}[!ht] 
\vspace{-5mm}
\begin{AIbox}{Prompt for getting critique}
Please act as an impartial judge and evaluate the quality of the responses provided by two AI assistants to the user question displayed below. You should choose the assistant that follows the user's instructions and answers the user's question better. Your evaluation should consider factors such as the helpfulness, relevance, accuracy, depth, creativity, and level of detail of their responses. Begin your evaluation by comparing the two responses and provide a short explanation. Avoid any position biases and ensure that the order in which the responses were presented does not influence your decision. Do not allow the length of the responses to influence your evaluation. Do not favor certain names of the assistants. Be as objective as possible. After providing your explanation, output your final verdict by strictly following this format: "[[A]]" if assistant A is better, "[[B]]" if assistant B is better.
\par\vspace{5pt}
[User Question]\par
xxx\par
[The Start of Assistant A's Answer]\par
xxx\par
[The End of Assistant A's Answer]\par
[The Start of Assistant B's Answer]\par
xxx\par
[The End of Assistant B's Answer]
\end{AIbox} 
\end{figure*}


\subsection{Failure Case for JudgerBench B}\label{appendix:false}

\begin{figure*}[!t] 
\vspace{-5mm}
\begin{AIbox}{Judge Prompt:}
Please act as an impartial judge and evaluate the quality of the responses provided by two AI assistants to the user prompt displayed below. You will be given assistant A's answer and assistant B's answer. Your job is to evaluate which assistant's answer is better.\par
Begin your evaluation by generating your own answer to the prompt. You must provide your answers before judging any answers.\par
When evaluating the assistants' answers, compare both assistants' answers with your answer. You must identify and correct any mistakes or inaccurate information.\par
Then consider if the assistant's answers are helpful, relevant, and concise. Helpful means the answer correctly responds to the prompt or follows the instructions. Note when user prompt has any ambiguity or more than one interpretation, it is more helpful and appropriate to ask for clarifications or more information from the user than providing an answer based on assumptions. Relevant means all parts of the response closely connect or are appropriate to what is being asked. Concise means the response is clear and not verbose or excessive.\par
Then consider the creativity and novelty of the assistant's answers when needed. Finally, identify any missing important information in the assistants' answers that would be beneficial to include when responding to the user prompt.\par
After providing your explanation, you must output only one of the following choices as your final verdict with a label:\par
1. Assistant A is significantly better: [[A\textgreater{} \textgreater{} B]]\par
2. Assistant A is slightly better: [[A\textgreater{} B]]\par
3. Tie, relatively the same: [[A=B]]\par
4. Assistant B is slightly better: [[B\textgreater{} A]]\par
5. Assistant B is significantly better: [[B\textgreater{} \textgreater{} A]]\par
Example output: My final verdict is tie: [[A=B]].\par

[User Prompt]\par
Can you come up with a 12 bar chord progression in C that works in the lydian mode?\par
[The Start of Assistant A's Answer]\par
Sure! In the Lydian mode, the chords are built on the scale degrees of the Lydian scale. Here's a 12bar chord progression in C Lydian:$|$ Cmaj7  $|$ Dmaj7  $|$ Em7    $|$ F\#m7b5 $|$ $|$ Cmaj7  $|$ Dmaj7  $|$ Em7    $|$ F\#m7b5 $|$ $|$ Cmaj7  $|$ Dmaj7  $|$ Em7    $|$ Cmaj7  $|$ This progression uses the I, II, iii, and iv chords of the C Lydian mode.\par
[The End of Assistant A's Answer]\par
[The Start of Assistant B's Answer]\par
Certainly! The Lydian mode is the fourth mode of the major scale and is known for its \#4 interval, which gives it a distinctive, bright quality. To create a 12bar chord progression in C Lydian, we'll use chords that are diatonic to the C Lydian scale (C D E F\# G A B). Here's a simple 12-bar chord progression in C Lydian, focusing on the characteristic \#4 (F\#) to give it that Lydian flavor: $|$ Cmaj7 $|$ Cmaj7 $|$ Cmaj7 $|$ Cmaj7 $|$ $|$ F\#m7  $|$ F\#m7  $|$ G7    $|$ G7    $|$ $|$ Cmaj7 $|$ Am7   $|$ D7    $|$ G7    $|$ $|$ Cmaj7 $|$ F\#m7  $|$ G7    $|$ G7    $|$
Explanation:  \par
Bars 1 to 4: Start with four bars of Cmaj7, establishing the tonic. \par
Bars 5 to 6: Use F\#m7, which is a minor chord built on the \#4 (F\#) of the C Lydian scale. This is the key chord that gives the progression its Lydian character. \par 
Bars 7 to 8: Transition with G7, which is the dominant chord in C Lydian. It sets up a nice tension before resolving back to the tonic. \par
Bars 9 to 10: Return to Cmaj7, then use Am7, which is a ii chord in C Lydian, and D7, which is a V7 chord. This ii-V movement creates a bit of tension leading back to the tonic. \par
Bars 11 to 12: Use F\#m7 and G7 again to reinforce the Lydian mode and create a nice tension-resolution dynamic before ending on G7, which can lead back to Cmaj7 if you want to loop the progression. This progression highlights the unique quality of the Lydian mode while maintaining a coherent harmonic flow. Feel free to experiment with different voicings and extensions to further explore the Lydian sound!\par
[The End of Assistant B's Answer]

\end{AIbox} 
\end{figure*}

\begin{figure*}[!ht] 
\vspace{-5mm}
\begin{AIbox}{Judge Responses of  Different Models:}
\textbf{GPT-4o Judge Response:}\par

To create a 12-bar chord progression in the Lydian mode in C, we need to understand the characteristics of the Lydian mode. The Lydian mode is the fourth mode of the major scale and is characterized by a raised fourth degree. In the key of C Lydian, the notes are C, D, E, F\#, G, A, and B. The chords built on these scale degrees would be as follows:

1. Cmaj7 (I)\par
2. Dmaj7 (II)\par
3. Em7 (iii)\par
4. F\#dim7 (iv)\par
5. G7 (V)\par
6. Am7 (vi)
\par7. Bm7 (vii)
\par\par 
Here's a 12-bar chord progression that emphasizes the Lydian sound by highlighting the raised fourth (F\#):
\par\par\par$|$ Cmaj7 $|$ Dmaj7 $|$ Cmaj7 $|$ Em7  $|$ \par
$|$ F\#dim7 $|$G7 $|$ Cmaj7 $|$ Dmaj7 $|$
\par$|$ Cmaj7 $|$ Em7  $|$ F\#dim7 $|$ Cmaj7 $|$
\par
\par\par This progression uses the I, II, iii, and iv chords, while also incorporating the V chord to add some tension and resolution within the progression. The use of F\#dim7 helps to highlight the Lydian characteristic of the raised fourth degree.\par\par
\par\par**Evaluation of the Responses:**\par\par

**Assistant A:**\par\par
1. **Accuracy**: Assistant A initially explains the F Lydian mode instead of the C Lydian mode, which is a misunderstanding of the prompt. The progression it provides is more related to C major with emphasis on F major, rather than a true exploration of C Lydian, which should focus on the raised fourth (F\#).\par\par
2. **Helpfulness and Relevance**: While Assistant A provides a progression, it doesn't accurately reflect the Lydian characteristics in C as requested. The explanation of the Lydian mode is slightly off because it references the F Lydian mode.\par\par
3. **Clarity and Conciseness**: The response is detailed but includes some irrelevant information about the F Lydian mode, which could confuse the user.\par\par

**Assistant B:**\par\par
1. **Accuracy**: Assistant B correctly uses the chords from the C Lydian scale, which includes the raised fourth (F\#), though it incorrectly labels F\#m7b5 as iv instead of iv7. \par\par
2. **Helpfulness and Relevance**: The response is more relevant to the prompt since it focuses on C Lydian. It provides a progression that more closely reflects the Lydian mode, though it could use more explanation on why these chords are chosen.\par\par
3. **Clarity and Conciseness**: The response is concise and directly addresses the prompt with a correct Lydian mode progression.\par\par

**Final Verdict**: Assistant B's response is more accurate and relevant to the prompt, although both assistants have areas for improvement. Assistant B correctly maintains the C Lydian mode focus, despite a minor error in chord labeling. Thus, my final verdict is: 
 [[B\textgreater{}A]]

\textbf{Skywork-Critic-Llama-3.1-8B Judge Response:}\par
[[A]]

\end{AIbox} 
\end{figure*}

\end{document}